\documentclass{article}
\usepackage{graphicx}
\usepackage{skak}

\title{Unanswerable Questions about Images and Texts}

\author{
Ernest Davis \\
Dept. of Computer Science \\ New York University \\ New York, NY 10012 \\
{\small davise@cs.nyu.edu}}

\newcommand{\te}{\mbox{$\exists$}}

\newcommand{\la}{\langle}
\newcommand{\ra}{\rangle}

\begin{document}
\maketitle

\begin{abstract}
Questions about a text or an image that cannot be answered raise distinctive
issues for an AI. This note discusses the problem of unanswerable questions 
in VQA (visual question answering), in QA (visual question answering), and
in AI generally.
\end{abstract}

The paper under discussion (Kafle, Shrestha, and Kanan 2019)
gives an extensive, detailed, 
and thoughtful review of the many
issues that have to be faced in designing datasets to train and evaluate
vision and language (V\&L) systems, and avoid ``Clever Hans'' effects
(Heinzerling  2019). It further
proposes a number of strategies to address the issues and mitigate the
problems.

Some of these problems are not specific to V\&L systems, but are instances
of problems that are common throughout machine learning (ML)-based AI systems.
In  particular,
Kafle, Shrestha, and Kanan write:
\begin{quote}
Human beings can provide explanations, point to evidence, and convey 
confidence in their predictions. They also have the ability to say ``I
do not know'' when the information provided is insufficient. However,
almost none of the existing V\&L algorithms are equipped with these abilities,
making their models highly uninterpretable and unreliable.

In VQA [visual question answering], 
algorithms provide high-confidence answers even when the question
is nonsensical for a given image, e.g. ``What color is the horse?'' for
an image that does not contain a horse can yield ``brown'' with a very
high confidence.
\end{quote}

There has been a flurry of recent papers addressing the issue of unanswerable 
questions, both in visual question answering (VQA) and in text-based question 
answering (QA). As far as can be judged from citations, these two directions
of research have been largely independent; that is, the VQA papers rarely 
cite QA research and vice versa. 
This note discusses the problem of unanswerable
questions in VQA (section~\ref{secVQA}) and in QA (section~\ref{secQA}) 
and compares them (section~\ref{secComparison}); 
they have some commonalities and many important
differences. 
Section~\ref{secRelatedWork} briefly discusses some related work from
knowledge-based AI.
Section~\ref{secGeneral} discusses the problem of unsolvable
problems and unanswerable questions in AI generally.
Section~\ref{secImportant} discusses why it is important for AI systems
to deal with unanswerable questions.

\section{Unanswerable questions in images}
\label{secVQA}
In the last few years many VQA datasets have been created in the computer
vision research community (see Kafle, Shrestha, and Kanan 2019 for a review).
For the most part, these datasets contain 
high-quality images that have been posted
to web sites, paired with questions 
that have been constructed, either automatically,
or by crowd workers, or by in-house participants 
(see Gurari et al., 2018, table 1). 
In most of these datasets, all of the questions are relevant to the image and
answerable from the image. As a result, as stated 
in the above quote from Kafle, Shrestha, and Kanan, a
system that has been trained on such a dataset will {\em always} output
an answer, even if the question is unanswerable, because in every example
in its training set, the proper response was to output an answer.
Researchers interested in detecting irrelevant and unanswerable questions have
therefore had to add these deliberately to their datasets.

Ray et al (2016) constructed irrelevant questions 
by randomly pairing an image from the VQA 
dataset\footnote{Regrettably, the acronym VQA refers both to the general
area of research and to this specific, widely used, dataset.}  
(Antol et al.  2015) with
a question from the same dataset; this technique, of course, mostly generates
questions that are entirely unrelated to the associated image.

\begin{figure}
\begin{center}
\includegraphics[width=6in]{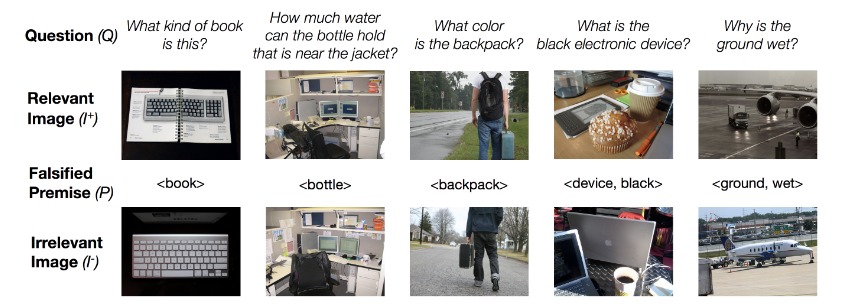}
\end{center}
\caption{Answerable and Unanswerable Questions in (Mahendru et al 2017).
The images are part of the VQA data set http://visualqa.org/
where they are published under a Creative Commons Attribution 
4.0 International License}
\label{figMahendru}
\end{figure}

Mahendru et al. (2017) use a more sophisticated
technique. They take an image/question pair $\la I^{+},Q \ra$ from 
the VQA dataset. They
then use simple NLP techniques 
to extract a premise $P$ from question $Q$; for instance, from the question
``Why is the ground wet?'' they extract the premise $\la$ground, wet$\ra$.
They then search for an image $I^{-}$ in the dataset that 
(a) is visually similar
to $I^{+}$; (b) violates one aspect of the premise; e.g. an image showing
dry ground.
The two pairs $\la I^{+},Q \ra$ and $\la I^{-},Q \ra$ 
are then added to the database; the discrimination that the first is answerable
and the second is unanswerable is generally quite challenging for VQA systems
(figure~\ref{figMahendru}).
Mahendru et al.
also use the unsatisfied premise to train a system that can give responses
that deny the premise e.g. ``The ground is not wet.''

\begin{figure}
\begin{center}
\includegraphics[width=6in]{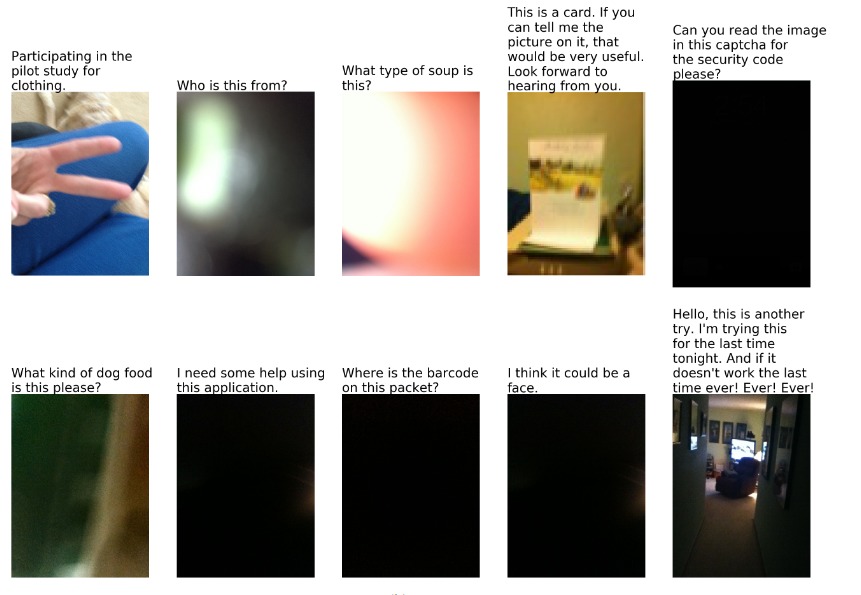}
\end{center}
\caption{Unanswerable questions in VizWiz. From (Gurari et al., 2018).
The images and texts are part of the VizWiz data set 
https://vizwiz.org/tasks-and-datasets/vqa/
where they are published under a Creative Commons Attribution 
4.0 International License}
\label{figVizWiz}
\end{figure}

The VizWiz dataset (Gurari et al. 2018) was constructed in an entirely 
different way from other VQA datasets. VizWiz is an app that 
allows blind or visually impaired users to take a picture and ask a question
about it, which is then answered by human employees or volunteers. With
the permission of the users, Gurari et al. collected
31,173 image/questions pairs, after
carefully screening by experts to eliminate images that might
have personal or embarrassing information. Crowdworkers were then used to
write new answers to the questions, or to mark the questions as unanswerable 
(figure~\ref{figVizWiz}).

The VizWiz dataset has markedly different characteristics from datasets such
as VQA:
\begin{itemize}
\item The images and questions reflect the practical needs of visually 
impaired users. This, obviously, yields a very different distribution of 
images and questions
than you get when you download images from the web and  then ask people
to pose interesting questions about them.
\item A significant fraction of the images are of very low quality, due to
the users' inability to judge.
\item Since the questions are from recorded speech, they are more informal
and conversational than the questions in VQA. A significant fraction are 
missing the first word or two, because the recording started late. Some of
them are not questions at all, but other kinds of expressions.
\end{itemize}

As a result, 28\% of the questions in VizWiz are unanswerable.

Other studies that have addressed the issues of unanswerable visual questions
include (Toor, Wechsler, and Nappi, 2018) and (Bhattacharya, Li, and
Gurari, 2019).

\begin{figure}
\begin{center}
\includegraphics[width=6in]{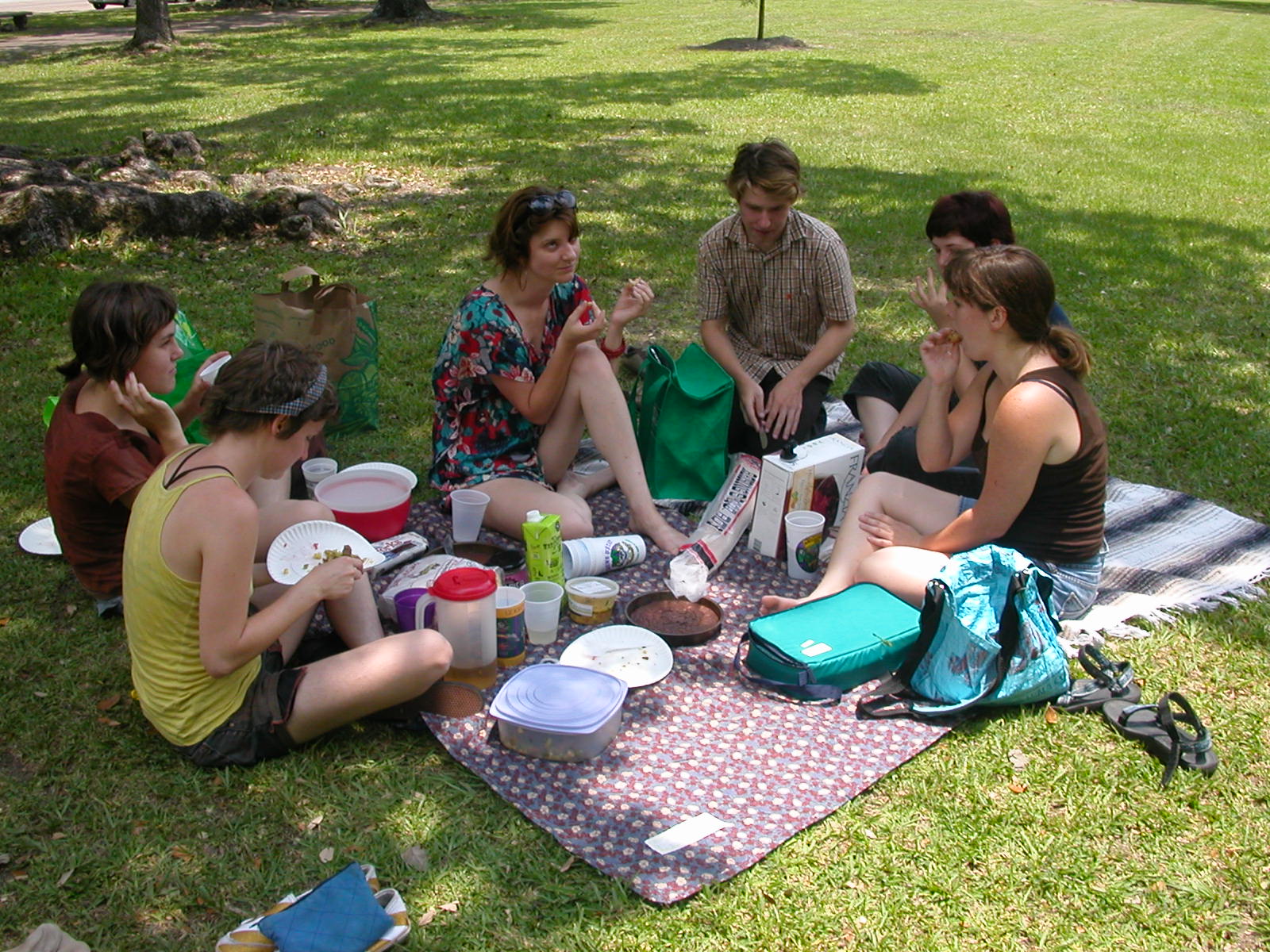}
\end{center}
\caption{Picnic. From Wikimedia Commons, with a Creative Commons License} 
\verb|https://commons.wikimedia.org/wiki/File:Our_pre-July_4th_picnic_NOLA.jpg|
\label{figPicnic}
\end{figure}

I have not found any dataset that collects visual questions asked by 
sighted users in a comparably natural setting.
Presumably, almost all of these would be questions that the asker 
cannot themselves easily answer just looking at the image; these are
thus ``unanswerable questions'' at least for this user in this circumstance. 
(There are some
exceptions; it is common for someone who is teaching
to ask a question for which they know
the answer.) For instance, 
looking at the picture of the picnic in figure~\ref{figPicnic}, 
one might naturally ask: Who are these people?  How do they know each other?
What did they have for lunch? Where and when did this take place? 
What were they talking about?  Did they have a good time?

\subsection{Categories of unanswerable questions about images}
\label{secCategories}

Unanswerable questions in VQA can be categorized in
terms of the reason that they are unanswerable. (These categories derive
in part from the discussions in the papers cited above and are in part 
original. In particular, Bhattacharya, Li, and Gurari, 2019 has a comparable
list of categories of questions for which crowdworkers have supplied a number
of different answers.) 
Where not otherwise specified, the questions below refer to
figure~\ref{figPicnic}.

{\bf V.1. Details that are hard or impossible to discern:} 
{\em Is the woman in the center wearing lipstick?
What colors are the eyes of the man in the 
center? What is the food on the plate of the woman with the 
yellow-green shirt?}

{\bf V.2. Information that is occluded:} {\em Is the leftmost woman in the brown 
shirt 
barefoot? Is there lemonade in the plastic pitcher with the red lid?}

{\bf V.3. Information that is out of the picture frame:} {\em What kind of 
tree is the
trunk visible in the top left corner? What is casting the shadow on the front
edge of the blanket?}

{\bf V.4. Indeterminate spatial relations:} {\em Is it more than  20 feet from the 
tree trunk in the top left to the vertical pole in the top center? Is the 
cardboard box touching the plastic cup that is in front of it? Who is
taller: The woman in the center or the woman on the far right?}

{\bf V.5. Information that is (probably) indicated in the photo, to a 
viewer with the appropriate general knowledge:}
{\em What is the material of the white 
rug with the black stripes? What decade was the photo taken?}

{\bf V.6. Information of a kind that is indicated in some photos but 
doesn't happen to be in this photo:} 
{\em What are the people eating for lunch?
Who is the youngest person at the picnic?}

{\bf V.7. Information that can rarely be indicated in any photo:} 
{\em Does the woman in the center have perfect pitch?
What is the conversation about?} 

{\bf V.8. Questions with a false premise:}
{\em What color is the horse? Which woman is punching the man? What colors are 
the boots?}

{\bf V.9. Low-quality image:} As mentioned, these are common in VizWiz, though
essentially non-existent in datasets such as VQA.

{\bf V.10. Completely irrelevant:} {\em What is the capital of Alaska?}  

{\bf V.11. Not a question:} {\em Hello, this is another try.} These exist in
some number in VizWiz.

{\bf V.12. Nonsense (i.e. meaningless in any context):} 
What did the rug eat for lunch?  Will the shadows set fire to the tree?

{\bf V.13. Gibberish:} {\em What color are the margialigoelntest 
farbitlangefs?}

Some of the questions that I have put in categories V.1-V.4 and V.6 may well be 
in category V.5; that is, a sufficiently knowledgeable or perceptive viewer
could answer them from the photo, though I cannot.
There may well be people who can tell whether the woman in the center is
wearing lipstick,
who can identify the tree at the top just from the trunk, or who can
judge accurately the distance from the tree to the pole, or who can tell 
whether the woman in brown is barefoot from the way she is sitting.
But some of the questions 
are unquestionably
impossible to answer; there is simply no way to tell whether
the woman has perfect pitch or what the conversation is about. (Of course,
any question at all is answerable if you have the necessary knowledge about
the individuals; if you recognize the woman in the center and you happen to 
know that she has perfect pitch, then you can answer the question, ``Does
the woman in the center have perfect pitch?'')

For categories V.1-V.7, the best answer would be ``I don't think it's possible to
tell'' when that is the case, or ``I can't judge'' when I think that perhaps
someone else could answer the question.
In category V.8, it would be better to deny the premise 
--- ``There's no horse in the picture,'' ``No one is punching the man'' 
--- and
better yet, when possible, to clear up the questioner's inferred confusion 
``Those aren't boots, they're black sandals.'' 
As discussed above,
the system developed by Mahendru et al. (2017) generates answers
that deny the premise.
For category V.9 the proper response is, ``There is a problem with the 
picture.''
For category V.10-V.13, the 
proper answer is ``What are you talking about?'' or ``Are you OK?'' AIs of
general intelligence should likewise be able to give these kinds of answers.

\section{Unanswerable questions about text}
\label{secQA}

A number of recent papers, stemming from the seminal paper (Rajpurkar, 
Jia, and Liang, 2018), have studied the use of unanswerable questions
about text as adversarial examples
to probe the depth of understanding in QA systems.
Rajpurkar, Jia, and Liang collected 53,775
unanswerable questions, using the following procedure:
Crowdworkers were shown a 25-paragraph article drawn from SQuAD,
an earlier database for text QA (Rajpurkar et al. 2016). 
They were asked to create, for each paragraph in the article,
between one and five questions that
were about the topic of the paragraph and referenced some entity in the
paragraph, but in fact were unanswerable from the paragraph. (They were
given an example as illustration.) The questions thus collected 
were added to SQuAD creating a new dataset SQuAD 2.0 (originally called
SQuADRUn).

Rajpurkar, Jia, and Liang tested a number
of high-quality QA systems on SQuAD 2.0, and found that they performed
significantly worse than on SQuAD; for instance, the F1 score on the 
best-performing system DocQA+ELMO dropped from 85.8\% on SQuAD to 66.3\% on
SQuAD 2.0 (human performance was 89.5\%). 

QA datasets prior to SQuAD 2.0 had contained a fraction of questions that were 
unanswerable, due mostly to noise. These were mostly easy to detect 
as they did not have plausible answers. Jia and Liang (2017) had
proposed a
rule-based method for generating unanswerable questions; these were less
diverse and less effective than the crowd-sourced questions created for
SQuAD 2.0.

A number of papers since have taken up the challenge posed by SQuAD 2.0.
The U-Net system (Sun, Li, Qiu, and Liu, 2018) uses three components ---
an ``answer pointers'', a ``no-answer pointer'', and an ``answer verifier'' 
--- and achieved an F-score of 75\% on SQuAD 2.0. Hu et al. (2019) likewise
used a multi-component system and achieved an F-score of 74.2\%.
Zhu et al. (2019) used
a pair-to-sequence model to generate their own corpus of unanswerable 
questions, and by training a BERT-based model on these, achieved an
F-score of 83\% over SQuAD 2.0. 

Rajpurkar, Jia, and Liang (2018) also manually analyzed a small subcollection of
the new unanswerable questions, and found that they fell into the categories
shown in table~\ref{tabSquadrun}.

\begin{table}
\begin{center}
\begin{tabular} {|lllr|}
\hline
 Reasoning & Description & Example & \% \\ \hline
&  & Sentence: {\small {\em ``Several hospital pharmacies have decided to}} & \\
T.1. Negation  & Negation word inserted & {\small {\em outsource high 
risk preparations \ldots ''}} & 9\% \\
& or removed. & Question: {\small {\em ``What types of pharmacy functions have 
{\bf never}}}
& \\
& & {\small {\em been outsourced?''}} & \\ \hline

& & S: {\small {\em ``the extinction of the dinosaurs \ldots allows the}} & \\
T.2. Antonym & Antonym used & {\small {\em tropical rainforest to spread out 
across the continent}} & 20\% \\
& & Q: {\small {\em ``The extinction of what lead to the {\bf decline} of}} & \\
& & {\small {\em rainforests?''}} & \\ \hline

& Entity, number, or date & S: {\small {\em ``These values are much 
greater than the 9-88 cm}}  & \\
T.3. Entity Swap & replaced with other & 
{\small {\em as projected \ldots in its Third Assessment Report.}} & 21\% \\
& entity, number, or date . & 
Q: {\small {\em ``What was the projection of sea level
increases in the}} & \\
& & {\small {\em {\bf fourth assessment report}?''}} & \\ \hline

& Word or phrase is & S: {\small {\em 
``BSkyB \ldots waived the charge for subscribers whose}} &  \\
T.4. Mutual & mutually exclusive & 
{\small {\em package included two or more premium channels.''}} & 16\% \\
Exclusion & with something for which & Q: {\small {\em ``Which service did BSkyB
{\bf give away for free}}} & \\
& an answer is present & {\small {\em {\bf unconditionally}?''}} & \\ \hline

& & S: {\small {\em ``Union forces left Jacksonville and confronted}} & \\
& Asks for condition that & {\small {\em a Confederate Army at the Battle of 
Olustee.}} & \\
T.5. Impossible & is not satisfied by & {\small {\em Union forces then 
retreated to Jacksonville}} & 4\%. \\
Condition & anything in the paragraph & {\small {\em and held the city for the 
remainder of the war.''}} & \\
& & Q: {\small {\em ``After what battle did Union forces leave}} & \\
& & {\small {\em Jacksonville {\bf for good}?''}} & \\ \hline
 
T.6. Other & Other cases where the & S: 
{\small }{\em ``Schuenemann et al. concluded in 2011 that the} & \\
Neutral & paragraph does not imply & 
{\small {\em Black Death \ldots was caused by
a variant of Y. pestis.''}} & 24\% \\
& any answer. & Q: {\small {\em ``Who discovered Y. pestis?''}} & \\ \hline

T.7. Answerable & Question is answerable & & 7\% \\
& (i.e. dataset noise) & & \\ \hline
\end{tabular}
\end{center}
\caption{Categorization of unanswerable questions in SQuAD 2.0. From
(Rajpurkar, Jia, and Liang, 2018).}
\label{tabSquadrun}
\end{table}

\section{Comparing categories of unanswerable questions}
\label{secComparison}

All the categories of textual unanswerable questions 
in table~\ref{tabSquadrun}, except T.7, are forms
of ``invalid premise'', and thus correspond to category V.8 of 
section~\ref{secCategories}; they categorize different strategies that
can be used to construct a question with an invalid premise from text. 
One could certainly find, or construct, unanswerable visual questions
corresponding to each of these. For instance, applying them to the
images in top row of figure~\ref{figMahendru}, we could formulate such 
questions as:
\begin{itemize}
\item[V.8.1] {\bf Negation:} Why isn't the ground wet?
\item[V.8.2] {\bf Antonym:} Why is the ground dry?
\item[V.8.3] {\bf Entity Swap:} What color is the shopping cart?
\item[V.8.4] {\bf Mutual Exclusion:} What is the pink electronic device?
\item[V.8.5] {\bf Impossible Condition:} What color shirt is the person
facing us wearing?
\end{itemize}
In the absence of a text, however, these categories are much less
distinct from one another.

The invalid premise questions constructed by Mahendru et al. (2017) 
include questions about objects not present in the image, such as ``What
color is the horse''. These are excluded from SQuAD 2.0 because their
verbal analogue is too easy.

Going in the other direction: Categories V.1-V.5 and V.9 have to do
with features of images not relevant to text.
Categories V.9-V.12 have to do with
low-quality data that is included in VizWiz but not in SQuAD or VQA.

Category V.7 ``Information
that can rarely be indicated in any photo'' has no analogue for text; if
a question has an answer, then there is almost always some way of expressing
that answer in language.

Category V.5, ``Information that is indicated in the photo, to a viewer
with appropriate general knowledge,'' applies to text no less than to 
images.  For instance, given the text
of example T.4,  ``Union forces left Jacksonville and confronted a 
Confederate Army at the Battle of Olustee,'' many of my 
American readers can answer the
question ``What war was the Battle of Olustee part of?'' By contrast, 
fewer American readers, seeing the text ``At the Battle of 
the White Mountain, the 
combined forces of Charles Bonaventure de Longueval and Johann 
Tserclaes defeated the army of Christian of Anhalt'', can immediately
answer the question ``What war was the Battle of White Mountain part of?'' But,
of course, there is no difference in kind here, only a difference in the
general knowledge of a particular audience. In this case, the results might
be quite different for a Czech audience.

I have not seen any discussion of this kind of question in either the QA or
VQA literature.  Presumably, the crowd workers rarely
generate these, either as answerable questions or as unanswerable
questions. AI researchers seeking to put together a reliable corpus for
training and testing would naturally tend to avoid these; they can hardly
be generated automatically in current technology, they are probably very
hard to recognize, they probably have poor interannotator agreement among
human judges. However, they are interesting test questions for systems
that try to integrate question answering with background knowledge.

\section{Other Related Work}
\label{secRelatedWork}
Standardized tests, of various kinds, have included 
multiple-choice questions, or yes/no questions, asking whether a specific
question can be answered from specified information. This is a regular
part of the GMAT exam for business school, known as
``data sufficiency'' problems. Such questions were part of
the math SATs at one date, but were eliminated because they were found
to be too susceptible to coaching (Chipman, 2005).\footnote{Interestingly, 
there were studies that suggested that female students did better than 
male students on these questions.}  I don't know whether problems of this kind
are included in any  reading comprehension tests.

The earliest reference that I know
to unanswerable questions in the AI literature is
an example attributed\footnote{I mentioned 
this myself in (Davis, 1990), but I did
not record there, and now do not remember, where I got it from --- perhaps
word of mouth --- and I cannot find any other mention of it Googling.}
to John McCarthy:
\begin{quote}
A. Is the President sitting down or standing up at this moment? \\
B. I haven't the faintest notion. \\
A. Think harder.
\end{quote}
The point being that B {\em knows\/} that thinking harder won't help, but
that it is not easy to say {\em how\/} he knows it.

Davis (1988, 1989) develops a logical theory, using a possible-worlds 
semantics in
which the limitations of perception due to occlusion or limited acuity
can be expressed and connected to a theory of knowledge. In principle, 
such theories could be part of a knowledge-based approach to identifying
unanswerable VQA questions.

\section{Unanswerable questions in AI generally}
\label{secGeneral}
Any AI system can be viewed as a question-answering system for a narrow
class of question. A machine translation system answers questions like ``How
do you express this English sentence in French?''  A chess program answers
the question ``What is the best move in this state of the game?'' 
Thus, some of the issues that arise with
unanswerable questions in QA or VQA reflect more general issues with 
unanswerable questions for AI programs generally.
What distinguishes QA systems is that the question is explicitly given 
in natural language, and that there is therefore a 
wide range of possible questions.  

AI systems generaly have a particular space of outputs considered useful,
and they typically generate an output in this output space 
no matter what the input. The AI
system may be entirely unsure that its answer is correct. It may have never
seen a similar example. There may be no good output for the particular
input; or the input may be meaningless; or the input may be random noise.
Nonetheless, AI systems mostly will confidently output what they consider
to be their best answer in this outpus space. When there is a problem with
the input, the chosen output 
often has more to do with the AI's
understanding of the output space than with the particular input.

In one widely reported incident a couple of years ago
(Greg, 2018), someone discovered that if you asked
Google Translate to translate, 
``dog dog dog dog dog dog dog dog dog dog
dog dog dog dog dog dog dog dog dog dog'' from Yoruba to English,
the result was 
\begin{quote}
Doomsday Clock is three minutes at twelve 
We are experiencing characters and a dramatic developments in the world, 
which indicate that we are 
increasingly approaching the end times and Jesus’ return
\end{quote}
(The omission of the period between ``twelve'' and ``We'' was in the
output from Google Translate.) This particular example no longer works; 
Google Translate output now just echoes the input.  But similar, 
if less entertaining, errors persist. For example, 
as of 1/17/2020, when asked to translate
``margialigoelntest farbitlangefs'' from English into Hebrew,
Google Translate outputs ``shodedei m'raglim rabim'', which means
``many spy bandits.'' 
If you ask Google Translate to translate it into Turkish, you get
``margialigoelntest Instagram Hesabındaki Resim ve Videoları farbitlangefs''
which means (I think)
``margialigoelntest the Instagram account photos and videos 
farbitlangefs''.\footnote{In fairness, Hebrew and Turkish are somewhat 
exceptional in this regards. In most languages Google Translate
simply echoes the input,
even in languages that don't use Roman characters, like Arabic and Russian;
it does not attempt to transliterate.}
As I will discuss below, in one form or another,
this kind of behavior is very easy for AI systems to fall into,
given the way that they are built.

\begin{figure}
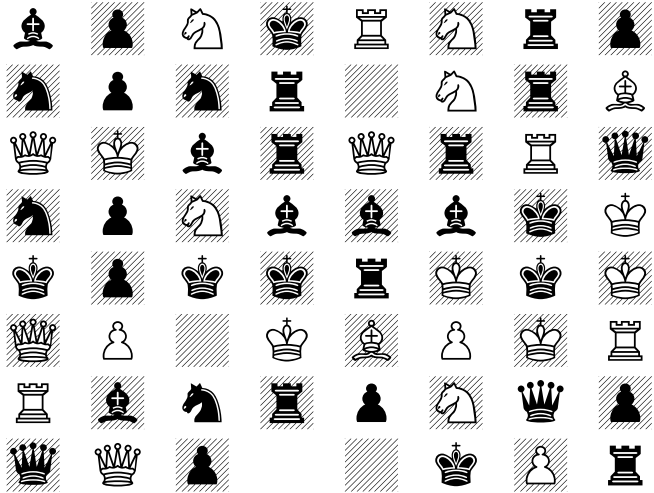

\begin{center}
 \begin{tabular} {cccccccc}
\BlackBishopOnWhite & \BlackPawnOnBlack & \WhiteKnightOnWhite &
\BlackKingOnBlack & \WhiteRookOnWhite & \WhiteKnightOnBlack &
\BlackRookOnWhite & \BlackPawnOnBlack \\
\BlackKnightOnBlack & \BlackPawnOnWhite & \BlackKnightOnBlack & 
\BlackRookOnWhite & \BlackEmptySquare & \WhiteKnightOnWhite &
\BlackRookOnBlack & \WhiteBishopOnWhite \\
\WhiteQueenOnWhite & \WhiteKingOnBlack & \BlackBishopOnWhite &
\BlackRookOnBlack & \WhiteQueenOnWhite & \BlackRookOnBlack &
\WhiteRookOnWhite & \BlackQueenOnBlack \\
\BlackKnightOnBlack & \BlackPawnOnWhite & \WhiteKnightOnBlack & 
\BlackBishopOnWhite & \BlackBishopOnBlack & \BlackBishopOnWhite &
\BlackKingOnBlack & \WhiteKingOnWhite \\
\BlackKingOnWhite & \BlackPawnOnBlack & \BlackKingOnWhite &
\BlackKingOnBlack & \BlackRookOnWhite & \WhiteKingOnBlack &
\BlackKingOnWhite & \WhiteKingOnBlack \\
\WhiteQueenOnBlack & \WhitePawnOnWhite & \BlackEmptySquare & 
\WhiteKingOnWhite & \WhiteBishopOnBlack & \WhitePawnOnWhite &
\WhiteKingOnBlack & \WhiteRookOnWhite \\
\WhiteRookOnWhite & \BlackBishopOnBlack & \BlackKnightOnWhite &
\BlackRookOnBlack & \BlackPawnOnWhite & \WhiteKnightOnBlack &
\BlackQueenOnWhite & \BlackPawnOnBlack \\
\BlackQueenOnBlack & \WhiteQueenOnWhite & \BlackPawnOnBlack & 
\WhiteEmptySquare & \BlackEmptySquare & \BlackKingOnWhite &
\WhitePawnOnBlack & \BlackRookOnWhite \\
 \end{tabular}
\end{center}
\caption{Unanswerable question in chess: What is the best move for white?}
\label{figChess}
\end{figure}

There are even unanswerable questions in chess. A chess program, such as the
chess version of AlphaZero (Silver et al. 2017), 
will accept as input any of the 
$13^{64}$ arrangements of the twelve pieces and ``empty'' on the chess board
as a legitimate input, and for any of these, it will output a move, because its
output space is the class of moves. But the rules of chess are only defined
for boards in which both players have exactly one king. (If a player has
two or  more kings, do you have to put them all checkmate?  Or to put one
of them into checkmate? 
to take all but one and put the last into checkmate?  The rules do not say,
because the situation cannot arise.) Therefore, in a situation like 
figure~\ref{figChess}, the proper response is ``This problem is meaningless,''
but AlphaZero will output a move.\footnote{It may be noted that only a tiny
fraction of such arrangements of chess pieces have one king for each player.
Thus, for the vast majority of chess boards --- though, of course, not of chess
boards that ever occur in practice --- a human being gives a better answer than
the best chess-playing programs.}

\subsection{An abstract framework}
\label{secAbstract}
It will be useful to setting up a general, abstract framework in which to 
discuss these issues.

Generally speaking AI systems, and for that matter computer programs of
any kind for a particular task, the actual ultimate objective can be formulated
as follows.  There is a class $X$ of inputs that are ``reasonable'' problems 
for $Q$. There is a class $Y$ of possible outputs. The task defines a relation 
$Q(x,y)$  meaning ``$y$ is a good output [or an acceptable output, or the best 
possible output] on the task for input
$x$.'' We assume that for every $x \in  X$ there is at least one 
$y \in Y$ such that $Q(x,y)$.
Define $Y$ to be the set of all good outputs for reasonable inputs: 
$Y = \{ y \: | \: \te_{x \in X} \: Q(x,y) \}$.

We now consider an ML-based AI system for this task. The construction of the
system begins by the assembling a corpus $C$. Ideally $C$ should be a 
representative sample of $X$, but often because of the way it is constructed,
it is biased and omits large swaths of $X$. 
You randomly divide $C$ into the training set
$T$ and the test set $S$.
You then construct an architecture that takes
arguments of form $I$ and produces outputs of form $O$. $I$ is often much 
larger than $C$;  $O$ can be equal to $Y$, or it can be much larger, or 
much smaller. You then apply the machine learning function to the training
set $T$ and learn a function $\Phi_{T}(x,y)$ which is the system's 
judgment of how good $y$ is as an output for $x$. 
Now, at inference time, given any $x \in I$, the machine will output
$\mbox{argmax}_{y \in O} \Phi_{T}(x,y)$, the most suitable output. (We ignore
the difference between the ideal objective function and the answer that
the mechanism actually outputs, which is not important for the analysis here.)
You test the machine over
$S$; of course, this test can detect overfitting to $T$ but not biases in
$C$. 

A few examples:
\begin{itemize}
\item In a speech recognition system, $X$ is the set of all things that 
someone might plausibly say. $Y$ is the set of meaningful natural language
expression. $I$ is the set of all waveforms. $C$ is the corpus of speech
samples that have been collected.
\item In a chess-playing program $X$ is the set of all game states that
plausibly might come up in an actual game. $Y=O$ is the set of all moves.
$I$ is the set of all placements of chess pieces on a chess board, including
figure~\ref{figChess}.
$T$ is the set of games that the program has looked
at in training.
\item For an image categorization system based on ImageNet, 
$X$ is the set of all reasonable
images. $Y$ is the set of all categories of objects that appear in images.
$I$ is the set of all pixel arrays. $O$ is the set of ImageNet categories.
In this case,
$O$ is considerably smaller than $Y$; that is, there are images of
objects where a good answer is a category not in ImageNet.
\item For high-quality VQA, 
$X$ is the set of all pairs of a high-quality image together with
an answerable question about the image. $I$ is the set of all pairs of a pixel
array and a character string. $Y$ is the class of all reasonable answers to
questions about images. $O$ is whatever the space of the possible outputs of
the system --- some subset of the space of all character strings.
\item For textual QA, $X$ is the set of pairs of a text together with an
answerable question about the text. $Y$ is the set of plausible answers to
such questions. $I$ is the set of pairs of character strings.\footnote{In many
recent systems, there is a preliminary step of converting words into word
embeddings; so, arguably, $I$ would be the set of pairs of sequences of
word embeddings.} $O$ is the space of character strings.
\end{itemize}

Ideally, a system should respond to an input $x$ with either ``It's impossible
to tell'' or ``the question is meaningless'' if the question is unanswerable; 
that is, if $x \in I$ but $x \not \in X$. (Our framework here is not rich
enough to distinguish between these two answers.) The system should respond,
``I don't know'' in either of the following cases:

\begin{itemize}
\item The system has no confidence in any possible answer: $\phi_{T}(x,y)$
is small for all $y \in O$. (This may be, of course, because the true answer
isn't in the space $O$ of the answers that the system can produce.)
\item The input $x$ is so dissimilar to any part of the training set that the
there is no reason for confidence that $\phi_{T}$ approximates $Q$ at $x$. 
That is, you have no confidence that the learning method will do out-of-domain
generalization with any quality. As an example, one can consider inputting a
line drawing to a vision system that has been trained exclusively on 
photographs, or an audio of French speech to a voice-recognition system trained
on English.
\end{itemize}

It is {\em possible\/}, of course, that the second case reduces to the first 
--- that whenever the input is too far from the training set, that will
be properly registered in the function $\phi_{T}$ --- but one certainly
cannot assume that that will hold in any particular case.

\subsection{An extreme case}
A case where $X$ and $Y$ vary to an extreme degree from $I$ and $O$
is the system for symbolic integration developed by Lample and Charton 
(henceforth LC)
(2019) using seq2seq technology. (Davis, 2019 is an extended critique.)
LC takes an elementary function $f$ (i.e. a composition of the 
arithmetic functions, the trigonometric and exponential functions and
their inverses) as input and is supposed to produce as output the indefinite 
integral of $f$, likewise expressed as a symbolic elementary functions.
It was trained and tested over a corpus of elementary functions whose 
integral is also elementary and, over this corpus, achieved a higher
success rate than state-of-the-art systems for symbolic mathematics such
as Mathematica and Matlab.

The problem, though, is that, for the vast majority of elementary functions 
$f$ of any significant
complexity, the indefinite integral of $f$ is non-elementary. 
For all of these, LC would happily return
an elementary function as an output, and that answer would necessarily be 
wrong. And the class of ``elementary functions whose integral is also an
elementary function'' is by no means a mathematically natural one.

So in terms of our framework: $X$ here are the class of elementary
functions. $I=O$ is the class of combinations of elementary and arithmetic
functions, both well-formed and ill-formed. (The output of LC is 
occasionally an ill-formed expression and LC will accept an ill-formed
expression as input without protest.)
$Y$ is the class of
elementary functions union the output ``The integral of the input is 
non-elementary.'' $C$ is a corpus of elementary functions
whose integral {\em is\/} elementary. LC achieved
an almost perfect success rate over a test set drawn from $C$, but would
almost never succeed over a test set drawn from complex expressions
in $X$.

The unanswerability of a question like, ``What is the integral of sin(sin(x))?''
is in a different category from those we have discussed for QA and VQA. The
question is meaningful --- the function has an indefinite integral --- but
the answer is not expressible in the language of elementary functions.

\section{Why are unanswerable questions important?}
\label{secImportant}
Why is it important to deal with unanswerable questions or invalid 
inputs at all?
Why not just say, ``Garbage in, garbage out'' and leave it at that?

Sometimes, in fact, that's fine. There is no reason that the AlphaZero team
at DeepMind should spend any time getting the program to respond correctly
to the situation in figure~\ref{figChess}. They can be entirely confident
that the only way their system will ever be used will be in the context of an
actual chess game, and equally sure that this situation may never arise in
an actual chess game.

But often it is important to respond properly to unanswerable questions and 
invalid inputs, for a number of reasons.
First, once a program is deployed to the real world,
invalid inputs do often occur, and they have to be flagged.  
Serious conventional programs for general
use are always built to address the problem of invalid inputs; it's a
basic principle of software engineering. If a program
bombs out without explanation 
when it gets invalid input, that's not great, but it is much worse
if the program returns a plausible-looking answer, because then the user
has no warning that there is a problem. A compiler which, analogously to LC,
produced correct executable code when the source was syntactically correct,
but also produced random executable code when the source is syntactically
incorrect would be a disaster.

The same applies for AI applications.
It's a ridiculously far-fetched example, of course, but suppose 
that a public figure
were to tweet something like, ``Despite the constant negative press covfefe'', 
and you were tasked with translating that into a foreign language. There are
various options you might consider, and it's not clear what would be best,
but simply rationalizing it to 
a phrase that makes some sense would not be a good option.

Similar considerations
apply to answering unanswerable questions in QA and VQA. 
As we have seen, invalid inputs, both in image and in question, are fairly
common in VizWiz. More generally,
suppose that you had a powerful VQA system that could reliably answer 
legitimate questions, but also answered unanswerable question. Now, someone
maliciously or not, gives the system a photo of a tall, thin African-American
man shaking hands with some terrorist $X$, and asks ``What is Obama doing with
$X$''? The AI system answers ``Obama is shaking hands with $X$.'' The 
mis- or disinformation now gets posted on social media, with the imprimatur
of the AI system.

Second, when AI outputs are being used as guides to action, then it becomes
extremely important to know how reliable they are. You want to avoid deciding
on an expensive or risky action on the base of unreliable information. This
is all the more important, because it is often completely impossible to
anticipate how an output will be used. A passage is mistranslated, a person
in an image is misidentified, speech is mistranscribed; OCR misreads a
text; the mistaken information is uploaded to
Web, or posted on social media; 
and then there is no telling what anyone will do with it.

Third, unanswerable questions can be used as adversarial examples
that give a means of probing the depths of an AI
system's understanding. If a VQA system confidently answers ``brown'' to the
question ``What color in the horse in figure~\ref{figPicnic}?''  or states
that the woman in the brown shirt is barefoot, then that raises eyebrows.
This kind of probing is the primary justification for including unanswerable
questions put forward in most of the current literature in VQA and QA.

This is probably useful, but a couple of cautions should be noted, in both
directions. On the one hand, if a system has been trained to respond 
correctly to
unanswerable questions, that may indeed reflect 
a deeper understanding, or it may
just mean that the system has learned ``Clever Hans'' tricks for that category
as well. 
Conversely, even if a system answers ``brown'' to the question, ``What color
is the horse?'' or ``Yes'' to the question, ``Is the woman in the brown
shirt barefoot?'', that may not reflect any gap in 
the system's understanding of horses,
brown, or bare feet. The gap may simply be in its understanding of what is
best to do when confronted with an unanswerable question. (School children 
likewise often guess randomly when confronted with this kind of ``trick''
question, if they are not expecting it.)

Fourth, under some circumstances, including unanswerable questions in training
may improve performance even on other tasks. Mahendru et al. (2017) report
that including unanswerable questions in their training set
had the effect of improving performance on tasks involving compositional 
reasoning.

Fifth, it is important to keep unanswerable questions as a category in
mind when comparing human beings or conventional computer systems against
AI systems.  Human beings generally deal with unanswerable questions 
flexibly and effectively; conventional software is often written to
give useful output for invalid inputs; AI systems mostly avoid the issue 
altogether.  The
claim that ``Chess programs are always better at analyzing a chess board
than humans'' is inarguably mostly true but comes with a very small asterisk 
for the reasons discussed above;
the claim that ``Watson can answer Jeopardy-style questions better than
human competitors'' comes with a significant asterisk; the claim that ``Lample
and Charton's symbolic integrator outperforms Mathematica
at symbolic integration'' comes with a very large asterisk, and indeed is
more false than true (Davis, 2019).
 
Finally, identifying what information is missing and why it missing is the
first step toward taking action to obtain the information. If the
image in figure~\ref{figPicnic} were the current perception of a robot,
who for some reason needs the answers to the question; it could move
its camera, or move to a new viewing position, or move some occluding
barrier, or wait until the 
picnickers stand up, or ask politely. Which of these would work and would
be appropriate of course depends on the particular question and 
circumstance, and choosing the answer depends on a meta-level 
understanding of what kind of information can be gotten under
what circumstances.

The ability to recognize input as ill-formed or nonsensical is one aspect
of
understanding what the input means when it is well-formed and meaningful.
The ability to realize that a picture or text
does not provide a particular kind of information is part of understanding
what kind of information it does provide. 
The ability to realize that you do not have the experience, 
knowledge, information, or cognitive capacity to be sure of any answer
is an aspect of self-knowledge critical in reasoning, acting, and
learning.

\subsection*{Acknowledgements}
Thanks to Yuling Gu and Wei Peng for helpful feedback.

\subsection*{References}
Antol, Stanislaw, Aishwarya Agrawal, Jiasen Lu, Margaret Mitchell, 
Dhruv Batra, C. Lawrence Zitnick, and Devi Parikh. 2015. VQA: Visual 
question answering. {\em ICCV-2015} 2425-2433.

Bhattacharya, Nilavra, Qing Li, and Danna Gurari. 2019.
Why Does a Visual Question Have Different Answers?. 
{\em ICCV-2019}, 4271-4280. 

Chipman, Susan F. 2005. Research on the women and mathematics issue. 
In Ann Gallagher and James Kaufmann (eds.) 
{\em Gender differences in mathematics: An integrative psychological 
approach} Cambridge University Press, 1-24.

Clark, Christopher, Kenton Lee, Ming-Wei Chang, Tom Kwiatkowski,
Michael Collins, and Kristina Toutanova, 2019.
BoolQ: Exploring the Surprising Difficulty of Natural Yes/No Questions.
{\em Proc. NAACL-HTL 2019,} 2924-2936.

Cook, William, 2012.
{\em In Pursuit of the Traveling Salesman: Mathematics at the Limits
of Computation.} Princeton University Press.

Davis, Ernest, 1988. Inferring Ignorance from the Locality of Visual
Perception. {\em AAAI-88,} 786-790.

Davis, Ernest, 1989. Solutions to a Paradox of Perception with Limited
Acuity. {\em KR-89,} 79-82.

Davis, Ernest, 1990. {\em Representations of Commonsense Knowledge.}
Morgan Kaufmann.

Davis, Ernest. 2019. The Use of Deep Learning for Symbolic Integration: A
Review of (Lample and Charton, 2019). 
arXiv preprint arXiv:1912.05752

Greg, 2019.
Dog’s final judgement: Weird Google Translate glitch delivers an 
apocalyptic message. {\em Daily Grail}, July 16, 2018. \\
https://www.dailygrail.com/2018/07/dogs-final-judgement-weird-google-translate-glitch-delivers-an-apocalyptic-message/.

Gurari, Danna, Qing Li, Abigale J. Stangl, Anhong Guo, Chi Lin, Kristen 
Grauman, Jiebo Luo, and Jeffrey P. Bigham. VizWiz grand challenge: 
Answering visual questions from blind people. 
{\em IEEE CVPR-2018,} 3608-3617

Henizerling, Benjamin. 2019 
NLP's Clever Hans Moment has Arrived: A review of Timothy Niven and 
Hung-Yu Kao, 2019: 
Probing Neural Network Comprehension of Natural Language Arguments. 
Heinzerling blog.
https://bheinzerling.github.io/post/clever-hans/

Hu, Minghao, Furu Wei, Yuxing Peng, Zhen Huang, Nan Yang, and 
Dongsheng Li. 2019. Read + verify: Machine reading comprehension with 
unanswerable questions. {\em AAAI-19,} 6529-6537.

Jia, Robin, and Percy Liang. 2017, 
Adversarial examples for evaluating reading comprehension systems. 
arXiv preprint arXiv:1707.07328.

Kafle, Kushal, Robik Shrestha, and Christopher Kanan. 2019. 
Challenges and Prospects in Vision Language Research. 
{\em Frontiers in Artificial Intelligence: Language and Computation,}
December, 2019

Lample, Guillaume and Fran\c{c}ois Charton. 2019. Deep Learning for 
Symbolic Mathematics. {\em NeurIPS-2019} 
arXiv preprint arxiv:1912.01412

Mahendru, Aroma, Viraj Prabhu, Akrit Mohapatra, Dhruv Batra, and Stefan 
Lee. 2017. The promise of premise: Harnessing question premises 
in visual question answering. arXiv preprint arXiv:1705.00601 

Rajpurkar, Pranav, Jian Zhang, Konstantin Lopyrev, and Percy Liang.  2016.
Squad: 100,000+ questions for machine comprehension of text. 
arXiv preprint arXiv:1606.05250.

Rajpurkar, Pranav, Robin Jia, and Percy Liang. 2018. 
Know What You Don't Know: Unanswerable Questions for SQuAD.
{\em ACL-2018.}

Ray, Arijit, Gordon Christie, Mohit Bansal, Dhruv Batra, and Devi Parikh. 
2016. Question relevance in VQA: identifying non-visual and false-premise 
questions. arXiv preprint arXiv:1606.06622 (2016).

Silver, David et al. 2017. 
Mastering the game of Go without human knowledge. 
{\em Nature} {\bf 550}(7676): 354–359.

Tan, Chuanqi, Furu Wei, Qingyu Zhou, Nan Yang, Weifeng Lv, and Ming 
Zhou. 2018. I know there is no answer: modeling answer validation for 
machine reading comprehension. {\em CCF International Conference on 
Natural Language Processing and Chinese Computing.} Springer. pp. 85-97. 

Taylor, Matthew E., Cynthia Matuszek, Pace Reagan Smith, and Michael J. 
Witbrock. 2007. 
``Guiding Inference with Policy Search Reinforcement Learning.'' 
{\em FLAIRS-2007} 146-151. 

Toor, Andeep S., Harry Wechsler, and Michele Nappi. 2017. Question 
part relevance and editing for cooperative and context-aware VQA (c2vqa)." 
{\em Proceedings of the 15th International Workshop on Content-Based 
Multimedia Indexing} 1-6.

Zhu, Haichao, Li Dong, Furu Wei, Wenhui Wang, Bing Qin, and Ting Liu. 
2019. Learning to ask unanswerable questions for machine reading 
comprehension. arXiv preprint arXiv:1906.06045 (2019).
\end{document}